\pdfoutput=1

\documentclass[11pt]{article}

\usepackage[review]{acl}

\usepackage{times}
\usepackage{latexsym}

\usepackage[T1]{fontenc}

\usepackage[utf8]{inputenc}

\usepackage{microtype}

\usepackage{inconsolata}

\usepackage{graphicx}
\usepackage{booktabs}
\usepackage{amssymb}
\usepackage{utfsym}
\usepackage{amsmath}
\usepackage{mathrsfs}

\usepackage{makecell}
\usepackage{pifont}
\newcommand{\xmark}{\ding{55}\,\,}
\usepackage{multirow}
\usepackage{blindtext}
\usepackage{multicol}

\title{Logical Reasoning over Natural Language as \\Knowledge Representation: A Survey}

\author {
    Zonglin Yang\textsuperscript{\rm 1}
    Xinya Du\textsuperscript{\rm 2}
    Rui Mao\textsuperscript{\rm 1}
    Jinjie Ni\textsuperscript{\rm 1}
    Erik Cambria\textsuperscript{\rm 1} \\
    \textsuperscript{\rm 1} Nanyang Technological University \\
    \textsuperscript{\rm 2} University of Texas at Dallas \\
    {\tt \normalsize \{zonglin.yang,rui.mao,jinjie001,cambria\}@ntu.edu.sg} \\
    {\tt \normalsize xinya.du@utdallas.edu}
}

\begin{document}
\maketitle
\begin{abstract}
Logical reasoning is central to human cognition and intelligence.
It includes deductive, inductive, and abductive reasoning.
Past research of logical reasoning within AI uses formal language as knowledge representation and symbolic reasoners.
However, reasoning with formal language has proved
challenging~(e.g., brittleness and knowledge-acquisition bottleneck).
This paper provides a comprehensive overview on a new paradigm of logical reasoning, which uses natural language as knowledge representation and pretrained language models as reasoners, including philosophical definition and categorization of logical reasoning, advantages of the new paradigm, benchmarks and methods, challenges of the new paradigm, possible future directions, and relation to related NLP fields.
This new paradigm is promising since it not only alleviates many challenges of formal representation but also has advantages over end-to-end neural methods.
This survey focus on transformer-based LLMs explicitly working on deductive, inductive, and abductive reasoning over English representation.
\end{abstract}

\section{Introduction}
An argument consists of premise(s) and a conclusion.
Logical reasoning is a form of thinking in which premises and relations between premises are used in a rigorous manner to infer conclusions that are entailed~(or implied) by the premises and the relations~\citep{Nunes2012}. 
It consists of three reasoning types, namely deductive reasoning, inductive reasoning, and abductive reasoning~\citep{flach2000abductive}~(more illustration on the categorization can be found in \S\ref{sec:definition}).
It is important since the ability to reach logical conclusions on the basis of prior information is recognized as central to human cognition
and intelligence~\citep{goel2017reasoning}.

The past research of logical reasoning within AI uses formal language~(e.g., first-order logic) as knowledge representation and symbolic reasoners~\citep{DBLP:journals/jlp/MuggletonR94}.
This paradigm has resulted in impressive applications such as expert systems~\citep{DBLP:journals/jim/MetaxiotisAP02}.
However, building and reasoning over formal language have proved 
challenging~\citep{musen1988brittleness}, with representative disadvantages of brittleness~(an expert system fails as long as its knowledge base does not contain complete knowledge for a problem) and knowledge-acquisition bottleneck~(human experts are needed to encode their knowledge with formal representation).

\begin{figure}[t]
\centering
\resizebox{1.0\columnwidth}{!}{
\includegraphics[]{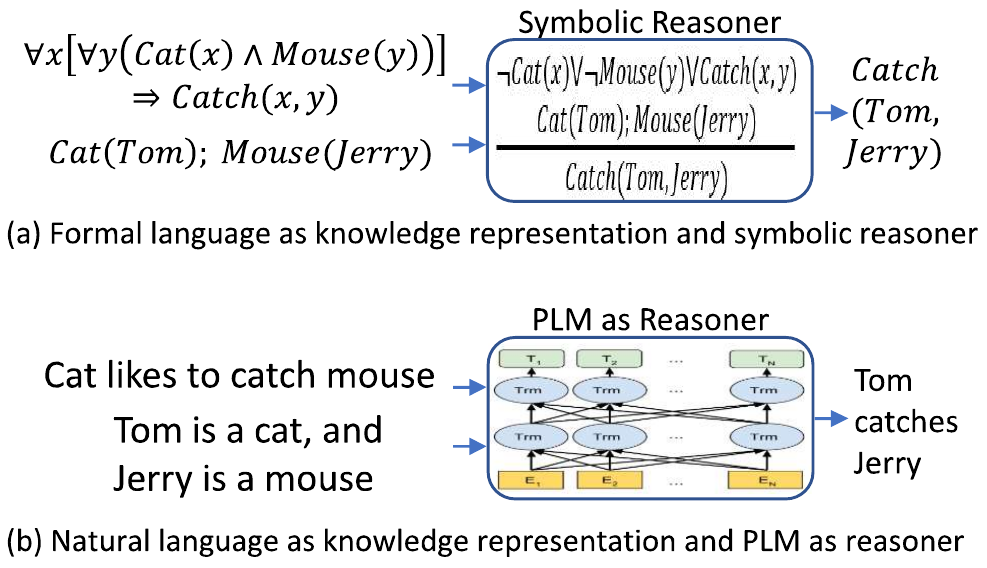}
}
\caption{Comparison between the previous paradigm which uses formal representation and symbolic reasoner, and the new paradigm which uses natural language as knowledge representation and LLM as reasoner. 
}
\label{fig:main_framework}
\end{figure}
Since the rapid development in language models, natural language has been explored as a new knowledge representation, and large language models~(LLMs) have been used as a new reasoner for deductive reasoning~\citep{DBLP:conf/ijcai/ClarkTR20}, abductive reasoning~\citep{DBLP:conf/iclr/BhagavatulaBMSH20}, and inductive reasoning~\citep{DBLP:journals/corr/abs-2212-10923}.
Therefore, all three reasoning types of logical reasoning have been investigated with natural language as knowledge representation.
This research also shows that LLMs can be finetuned or prompted to perform well for each of the reasoning types.

In this paper, we summarize the three previously separately investigated logical reasoning types together, referred as logical reasoning~(from the perspectives of deductive, inductive, and abductive reasoning) over natural language as knowledge representation and PLMs as reasoners~(LRNLP), and provide an in-depth survey of LRNLP.

Illustrated in Figure~\ref{fig:main_framework},
LRNLP means a new paradigm for logical reasoning that uses new knowledge representation~(natural language) and new reasoner~(LLM).
LRNLP can also be seen as a set of tasks on the three reasoning with a constraint of natural language representation and LLM reasoners.
The latest methods for the LRNLP tasks are generally modular: multiple LLMs each as one module playing a different function, combined together to perform complex tasks.
They make one step of reasoning
with one inference of LLM.
For complex problems, they usually have access to a knowledge base that stores relevant textual knowledge to be retrieved as premises to support the reasoning process to reach a conclusion, which might be used as a new premise for the next step's reasoning.
By iteratively repeating this process, a final conclusion may be made.
Although it looks similar to expert systems,
we discuss how LRNLP is possible to overcome many main challenges of the previous paradigm
such as brittleness 
in \S\ref{subsec:advantages_over_formal_language}.

In addition to the comparison with formal language, in \S\ref{subsec:advantages_over_nesy_methods} we discuss that LRNLP could be viewed as a new type of neural-symbolic method, which has unique advantages over existing neuro-symbolic methods.
We also discuss how LRNLP, as a neuro-symbolic method, has advantages over existing end-to-end neural methods~(e.g., explainability, controllability, less catastrophic forgetting) in \S\ref{subsec:advantages_over_e2e_neural_methods}.
These advantages make an LRNLP system possible to deal with many challenging problems.


In the remaining sections of this survey, we review papers on LRNLP~(including deductive reasoning~\S\ref{sec:deductive}, inductive reasoning~\S\ref{sec:inductive}, and abductive reasoning~\S\ref{sec:abductive}), and list challenges~(\S\ref{sec:challenges}.
%
Our main focus is to understand the language model's logical reasoning ability through the three sub-types of logical reasoning to provide finer analysis and avoid ambiguity on which type of reasoning it is conducting.
Therefore, we focus on papers that using transformer-based LLMs explicitly working on deductive, inductive, or abductive reasoning tasks.
These papers all adopt English as knowledge representation.
In \S\ref{appen:relation_to_related_fields} we discuss the relation of LRNLP to related NLP fields~(e.g., commonsense reasoning), which could help to form a clear shape of LRNLP in NLP.
%
For each reasoning sub-type, we summarize existing task formulations, datasets, and methods under each task.

\section{Definition and Categorization}
\label{sec:definition}
There are many subjects related to logical reasoning, including philosophy, logic, and AI.
Among them, the definition and categorization aspects of logical reasoning are handled by philosophy research. 
However, debate exists in philosophy research on the categorization of logical reasoning.
We leave a detailed description of the debate in philosophy research in \S\ref{appen:difference_inductive_abductive} and only leave the conclusions here according to philosophy research.

In general, logical reasoning consists of deductive, inductive, and abductive reasoning~\citep{console2000relations}.
Given an argument consisting of premises and a conclusion, we define the sub-type of logical reasoning it involves below:

{\it Definition for deductive reasoning}: the premises can conclusively provide support for the conclusion, i.e. if the premises are all true, it would be impossible for the conclusion to be false.

{\it Definition for inductive reasoning}: the premises cannot conclusively provide support for the conclusion, since the conclusion generalizes existing information in premises to new knowledge, which has a wider applicable scope than those in premises.

{\it Definition for abductive reasoning}: the premises cannot conclusively provide support for the conclusion, since the conclusion contains more specific information over the premises~(most commonly used as generating most probable explanations).

Please note that according to \citet{console2000relations}, inductive reasoning and abductive reasoning are not exclusive to each other.

\section{Advantages of LRNLP}
\label{sec:advantages_of_new_paradigm}

\subsection{Advantages over Formal Language}
\label{subsec:advantages_over_formal_language}

Building and reasoning over formal language have proved challenging~\citep{musen1988brittleness,DBLP:journals/ml/CropperDEM22}, with disadvantages such as (1) \textit{brittleness}~(expert system fails when its knowledge base does not contain complete knowledge for a problem), (2) \textit{knowledge-acquisition bottleneck}~(human experts are needed to encode their knowledge with formal representation),
(3) \textit{inability to handle raw data such as natural language}, (4) \textit{sensitivity to label errors}, and (5) \textit{failure to recognize different symbols with similar meanings}.

Nevertheless, the new paradigm of logical reasoning, LRNLP, has systematic strengths over these challenges.
Specifically, LLMs contain knowledge themselves~\citep{DBLP:conf/emnlp/DavisonFR19}, which makes it possible for them to provide good answers even when some required explicit knowledge is not present in a knowledge base~\citealp{talmor2020leap}~(less brittle), and be less affected by input errors~\citep{DBLP:conf/emnlp/0001ZHWZJ021}. 
In addition, with natural language as knowledge representation, such a system can naturally handle raw input,
and it is possible to utilize the enormous web corpora to automatically construct rule bases using information extraction~\citealp{DBLP:reference/db/Ji18}
(less affected by knowledge-acquisition bottleneck);
using embeddings for concepts~\citep{DBLP:conf/nips/MikolovSCCD13}, it semantically ``understands'' the meaning of symbols and therefore robust for paraphrasing. 

\subsection{Advantages over Neuro-symbolic Systems}
\label{subsec:advantages_over_nesy_methods}

LRNLP could be seen as a new type of neuro-symbolic in addition to the existing 6 types\cite{kautz2022third}, as its goal and design of methodology are typically symbolic~(logical reasoning with knowledge bases), while avoiding any symbolic representation, using~(currently pure) neural methods.
Therefore LRNLP can avoid many bottlenecks of the other neuro-symbolic methods caused by symbolic representation, such as symbolic knowledge acquisition and scalability~\citep{wang2022towards}.

\subsection{Advantages over E2E Neural Methods}
\label{subsec:advantages_over_e2e_neural_methods}
As a neuro-symbolic method, LRNLP systematically has some advantages over end-to-end neural methods, such as interpretability~\citealp{DBLP:journals/ipm/CambriaMMMN23}~(since it is ususally stepwise), more controllability~(LRNLP reasons following a given knowledge base), and less catastrophic forgetting~(LRNLP uses an explicit knowledge base).

\section{Deductive Reasoning}
\label{sec:deductive}

\subsection{Existing Task Formulations}
\begin{table}[]
\resizebox{1.0\columnwidth}{!}{
\begin{tabular}{cccccccc}
\toprule
Dataset           & \thead{Human\\ written} & Realistic & \thead{Multi-\\step} & \thead{Theory\\ included} & \thead{Theory\\ sufficient} & \thead{Proof\\ generation} & Size  \\ \midrule
D*                & \xmark            & \xmark        & \checkmark        & \checkmark             & \checkmark               & \xmark         & 500k  \\
ParaRules         & \checkmark           & \xmark        & \checkmark        & \checkmark             & \checkmark               & \xmark         & 40k   \\
Birds-electricity & \checkmark           & \checkmark       & \checkmark        & \checkmark             & \checkmark               & \xmark         & 5k    \\
Leap-of-thought   & \xmark            & \checkmark       & \xmark         & \checkmark             & \xmark                & \xmark         & 33k   \\
PARARULE-Plus     & \xmark             &   \xmark       & \checkmark    &        \checkmark          &    \checkmark          &     \xmark      &  400k     \\
FOLIO             & \checkmark           & \checkmark       & \checkmark        & \checkmark             & \checkmark               & \xmark         & 1,435 \\ \midrule
D*(CWA)           & \xmark            & \xmark        & \checkmark        & \checkmark             & \checkmark               & \checkmark        & 500k  \\
D*(OWA)           & \xmark            & \xmark        & \checkmark        & \checkmark             & \xmark                & \checkmark        & 500k  \\
EntailmentBank    & \checkmark           & \checkmark       & \checkmark        & \checkmark             & \checkmark               & \checkmark        & 1,840 \\
ENWN              & \checkmark           & \checkmark       & \checkmark        & \checkmark             & \checkmark               & \checkmark        & 100  \\ \bottomrule
\end{tabular}}
\caption{Summary of deductive reasoning datasets: D*, ParaRules \& birds-electricity~\citep{DBLP:conf/ijcai/ClarkTR20}; leap-of-thought~\citep{talmor2020leap};
PARARULE-Plus~\citep{bao2022multi}; FOLIO~\citep{DBLP:journals/corr/abs-2209-00840}; D*(CWA) \& D*(OWA)~\citealp{DBLP:conf/acl/TafjordDC21}; EntailmentBank~\citep{DBLP:conf/emnlp/DalviJTXSPC21}; ENWN~\citep{DBLP:journals/corr/abs-2211-00614}.
}
\label{tab:deductive_reasoning_datasets}
\end{table}

Existing tasks for deductive reasoning can be summarized as hypothesis classification, proof generation, proof generation with incomplete information, and implication enumeration.
Datasets for tasks are summarized in Table~\ref{tab:deductive_reasoning_datasets}.
``Proof generation'' tab with \xmark means it is for hypothesis classification task.

\paragraph{Hypothesis Classification}

Each data example for hypothesis classification task is a tuple $(theory, hypothesis, correctness)$, where $theory$ typically has the form $(fact^*, rule^*)$, $hypothesis$ is a question, and $correctness$ can be $True$ or $False$ (or $Unknown$).
This task requires to predict the $correctness$ for the $hypothesis$ given the $theory$.

\paragraph{Proof Generation}
The proof generation task has the same setting as the hypothesis classification task, except that in addition to predicting a $correctness$, the proof generation task also requires providing a $proof$ given $theory$ to explain the $correctness$.
The $proof$ is a directed tree $(\mathcal{N}, \mathcal{E})$ with nodes $n \in \mathcal{N}$ and edges $e \in \mathcal{E}$.
Each node is an item of knowledge in $theory$~(usually a $fact$ or a $rule$), or a generated intermediate reasoning conclusion, or the $hypothesis$ itself;
Each edge points from a premise node to a conclusion node to form a deductive argument, which typically needs one-step inference~(not multi-step).

\paragraph{Proof Generation with Incomplete Information}
This task is the same as the proof generation task, except that $theory$ lacks one $node$ to form a complete $proof$. 
Specifically, given $theory$, it requires to predict the $correctness$ of $hypothesis$ with a $proof$, as well as recovering the missing $node$.

\paragraph{Implication Enumeration}
Given a $theory$,
this task requires to enumerate implications
of the $theory$, using deductive reasoning.



\subsection{Methods}
\begin{table*}[]
\centering
\resizebox{1.91\columnwidth}{!}{
\begin{tabular}{ccccccccc}
\toprule
 Method   & \thead{Generation\\ based} & \thead{Inference w/\\ hypothesis} & Stepwise & \thead{Proof \\direction} & \thead{Heuristic\\ search} & Verifier & \thead{Human-authored\\ realistic proof} & Stage \\ \midrule
PRover~\citep{DBLP:conf/emnlp/SahaGSB20}      & \xmark       & \checkmark     & \xmark       & N/A     & N/A      & \xmark       & \xmark     & 1 \\
multiPRover~\citep{DBLP:conf/naacl/SahaYB21} & \xmark       & \checkmark     & \xmark       & N/A     & N/A      & \xmark       & \xmark     & 1 \\
EntailmentWriter~\citep{DBLP:conf/emnlp/DalviJTXSPC21}    & \checkmark      & \checkmark     & \xmark       & N/A     & N/A      & \xmark       & \checkmark    & 1 \\ \midrule
ProofWriter~\citep{DBLP:conf/acl/TafjordDC21} & \checkmark      & \xmark      & \checkmark      & $\rightarrow$ & \xmark       & \xmark       & \xmark     & 2 \\
EVR~\citep{DBLP:conf/naacl/LiangBS21} & \checkmark      & \xmark      & \checkmark      & $\leftarrow$& \xmark       & \xmark       & \xmark     & 2 \\
IBR~\citep{DBLP:conf/naacl/QuCGDX22} & \xmark       & \checkmark     & \checkmark      & $\leftarrow$& \checkmark      & \xmark       & \xmark     & 2 \\
IRGR~\citep{DBLP:conf/naacl/Ribeiro0MDWZCXH22}& \checkmark      & \checkmark     & \checkmark      & $\rightarrow$ & \checkmark      & \xmark       & \checkmark    & 2 \\
SI~\citep{DBLP:journals/corr/abs-2205-09712} & \checkmark      & \xmark      & \checkmark      & $\rightarrow$ & \checkmark      & \xmark       & \xmark     & 2 \\
FaiRR~\citep{DBLP:conf/acl/SanyalS022}       & \checkmark      & \xmark      & \checkmark      & $\rightarrow$ & \checkmark      & \xmark       & \xmark     & 2 \\
MetGen~\citep{DBLP:conf/naacl/HongZYZ22}      & \checkmark      & \xmark      & \checkmark      & Both    & \checkmark      & \xmark       & \checkmark    & 2 \\
SCSearch~\citep{DBLP:journals/corr/abs-2201-06028}    & \checkmark      & \xmark      & \checkmark      & $\rightarrow$ & \checkmark      & \xmark       & \checkmark    & 2 \\ \midrule
ADGV~\citep{DBLP:journals/corr/abs-2211-00614}& \checkmark      & \xmark      & \checkmark      & Both    & \checkmark      & \checkmark      & \checkmark    & 3 \\
NLProofS~\citep{DBLP:journals/corr/abs-2205-12443}    & \checkmark      & \checkmark     & \checkmark      & $\rightarrow$ & \checkmark      & \checkmark      & \checkmark    & 3 \\
Entailer~\citep{DBLP:journals/corr/abs-2210-12217}    & \checkmark      & \checkmark     & \checkmark      & $\leftarrow$& \checkmark      & \checkmark      & \checkmark    & 3 \\
Teachme~\citep{DBLP:journals/corr/abs-2204-13074}     & \checkmark      & \checkmark     & \checkmark      & $\leftarrow$& \checkmark      & \checkmark      & \xmark     & 3\\ \bottomrule
\end{tabular}}

\caption{
Methods for Proof Generation task.
``Generation based'' means whether $proof$ is created by generative inference model, otherwise is by utilizing embeddings to classify nodes and edges of $proof$.
``Inference w/ $hypothesis$'' means whether $hypothesis$ is provided during inference.
$\rightarrow$ and $\leftarrow$ denote forward/backward stepwise proof generation.
``Heuristic search'' with \xmark means exhaustive search.
``Human-authored realistic proof'' means whether the dataset adopted uses human-authored $proof$, whose contents are consistent with the real world.
}
\label{tab:proof_generation_methods}

\end{table*}

\subsubsection{Hypothesis Classification}
There are mainly three categories of methods for the hypothesis classification task.
The first category only conducts the classification task itself;
the second category can predict $correctness$ as well as generate a $proof$. 
However, the $correctness$ is not necessarily consistent with the predicted $proof$.
The third category is similar to the second, except that $correctness$ always follows $proof$.

Until now, methods from the first category directly use transformer-based LLMs~\citep{vaswani2017attention}, aiming at analyzing and benchmarking their performance.
%
Specifically,~\citet{DBLP:conf/ijcai/ClarkTR20} find that finetuned RoBERTa-large~\citep{DBLP:journals/corr/abs-1907-11692} can achieve 95\%+ accuracy on the test set of D* and ParaRules datasets;
\citet{talmor2020leap} further demonstrate that LLMs can be finetuned to reliably perform deductive reasoning using both implicit, pretrained knowledge and explicit natural language statements~($theory$) to make predictions;
\citet{DBLP:journals/corr/abs-2209-00840} evaluate finetuned medium-sized language models and few-shot prompting on LLMs on the FOLIO dataset.
However, they find that LLM with few-shot prompting only performs slightly better than random results.

The second category methods typically infer LLM only once, and then utilize the final layer embeddings or generations to obtain $correctness$ and $proof$.
Specifically, PRover~\citep{DBLP:conf/emnlp/SahaGSB20} and multiPRover~\citep{DBLP:conf/naacl/SahaYB21} use the [CLS] token to predict $correctness$, and leverage the final layer embeddings of knowledge items in $theory$ to generate $proof$;
All-At-Once ProofWriter~\citep{DBLP:conf/acl/TafjordDC21} and EntailmentWriter~\citep{DBLP:conf/emnlp/DalviJTXSPC21} generate $correctness$ and linearized $proof$ at the same time.

The third category methods create a $proof$ first, and then predict $correctness$ from the $proof$.
\S\ref{subsec:proof_generation} illustrates these methods in detail.




\subsubsection{Proof Generation}
\label{subsec:proof_generation}

Methods for this task are summarized in Table~\ref{tab:proof_generation_methods}.
Current methods for the proof generation task roughly consist of three stages. 
In each stage, one key new technique is considered and developed.
In stage 1, LLMs are used for forming $proof$ in one inference step.
In stage 2, modular-based, stepwise frameworks are developed to create $proof$~(each module is usually implemented with a single LLM).
In stage 3, a verifier is added as a new module to make sure that each reasoning step reflects the belief of LLMs.
We summarize the experiment results in \S\ref{appen:evaluation} and the model structures in \S\ref{appen:model_structure_pretraining_data_used}.


Methods for stage 1 typically utilize the last layer embeddings~\citep{DBLP:conf/emnlp/SahaGSB20,DBLP:conf/naacl/SahaYB21} or generations~\citep{DBLP:conf/acl/TafjordDC21,DBLP:conf/emnlp/DalviJTXSPC21} to create $proof$.
Methods utilizing embedding typically (1) obtain an averaged embedding for each knowledge item in $theory$, and (2) pass each embedding to a node classifier, and each embedding pairs to an edge classifier to predict nodes and edges for $proof$.
Constraints are usually used to enforce the structure of $proof$.
Generation methods directly generate linearized $correctness$ and full $proof$ given linearized $theory$ and $hypothesis$.



The motivations of stage 2 methods are generally concerned with end-to-end methods, which is considered to lack interpretability~\citep{DBLP:conf/naacl/LiangBS21,DBLP:conf/naacl/QuCGDX22,DBLP:conf/acl/SanyalS022,DBLP:journals/corr/abs-2201-06028}, suffer from compositional generalization problems~\citep{DBLP:conf/naacl/LiangBS21,DBLP:journals/corr/abs-2205-09712}, have limited input size~\citep{DBLP:conf/naacl/Ribeiro0MDWZCXH22}, are not causal~\citep{DBLP:journals/corr/abs-2205-09712}, and lack constraints on inference validity~\citep{DBLP:conf/naacl/HongZYZ22}.

Methods in stage 2 can be summarized as having two components, an inference module and a reasoning controller.
The inference module can be a deduction module~\citep{DBLP:conf/acl/TafjordDC21,DBLP:conf/naacl/Ribeiro0MDWZCXH22,DBLP:journals/corr/abs-2205-09712,DBLP:conf/acl/SanyalS022,DBLP:journals/corr/abs-2201-06028}, an abduction module~\citep{DBLP:conf/naacl/LiangBS21,DBLP:conf/naacl/QuCGDX22}, or both~\citep{DBLP:conf/naacl/HongZYZ22,DBLP:journals/corr/abs-2211-00614}.
The deduction module performs deductive reasoning, and reasons forwardly from $theory$ to $hypothesis$ to construct $proof$;
the abduction module performs abductive reasoning, and reasons backwardly from $hypothesis$ to $theory$ to construct $proof$.
The reasoning controller in general performs a search process that each step it searches through the $theory$ and generated intermediate conclusions space to select~(retrieve) premises for the next step inference.
The search processes include exhaustive search~\citep{DBLP:conf/acl/TafjordDC21,DBLP:conf/naacl/LiangBS21} or heuristic search~\citep{DBLP:conf/naacl/QuCGDX22,DBLP:conf/naacl/Ribeiro0MDWZCXH22,DBLP:journals/corr/abs-2205-09712,DBLP:conf/acl/SanyalS022,DBLP:journals/corr/abs-2201-06028,DBLP:conf/naacl/HongZYZ22,DBLP:journals/corr/abs-2211-00614}.
The reasoning controller usually can also stop the search process if it detects the goal.


Motivation of stage 3 methods is similar, basically that
stage~2 methods lack explicit verifiers to avoid hallucinating invalid steps~\citep{DBLP:journals/corr/abs-2205-12443}, and to ensure that the inference processes reflect LLM's own beliefs~\citep{DBLP:journals/corr/abs-2210-12217}.

Methods in stage 3 can be summarized as utilizing explicit verifier(s)~(implemented with a LLM) to check the validity of each inference step.
One way is to add a new module (additional to the inference module and reasoning controller in stage 2), working as a ``fact checker'' to verify the generated inference step~\citep{DBLP:journals/corr/abs-2205-12443,DBLP:journals/corr/abs-2210-12217};
The other one, called round-trip consistency, is only suitable for methods that use both deduction and abduction modules, where deduction and abduction modules work as the verifier for each other~\citep{DBLP:journals/corr/abs-2211-00614}.


In addition to the general 3 stages, a new aspect is attended to, which is whether teachable by humans.
Build based on Entailer~\citep{DBLP:journals/corr/abs-2210-12217}, TeachMe~\citep{DBLP:journals/corr/abs-2204-13074} shows that user corrections can help override erroneous model beliefs, and that a system can gradually improve by accumulating user corrections.
Compared to Entailer, it adds an interaction module and a dynamic memory module to obtain and store human corrections.

\subsubsection{Proof with Incomplete Information}

ADGV~\citep{DBLP:journals/corr/abs-2211-00614} is the only method focusing on this task.
It uses both deduction and abduction modules, and the reasoning controller performs heuristic search.
The abduction module is used to recover the missing premise.


\subsubsection{Implication Enumeration}
\citet{DBLP:conf/acl/TafjordDC21} is the only paper mentioned this task.
They compare the performance of ``All-At-Once'' and ``Iterative'' ProofWriter on this task.
They find that ``All-At-Once'' performs worse, mainly because it struggles with problems that are more complex than training examples.


\subsection{Robustness of LLM as Reasoner}
\label{subsec:robustness_deductive_reasoning}


The previously introduced methods only focus on solving the deductive reasoning tasks, while it is unclear whether LLMs can be used as robust deductive reasoners.
To investigate the problem, 
\citet{DBLP:conf/ijcai/GaskellMTS22} create a more challenging synthetic dataset on hypothesis classification task in terms of complexity, and test LLM's performance on it. 
They find that with large and complex enough training examples, transformers can perform well on the dataset.
In addition, they find that transformers exhibit some degree of generalization and scale-invariance ability;
\citet{DBLP:conf/aaai/0001S22} propose an adversarial attack method for synthetic datasets on the hypothesis classification task.
They find that transformers are often fooled if the query literally appears within the body of a rule, and transformers struggle to correctly bind variables on either side of a rule;
\citet{DBLP:journals/corr/abs-2205-12598} proposed a synthetic deductive reasoning dataset to evaluate the robustness of language models to minimal logical edits in the inputs and different logical equivalence conditions, and find that LLMs are not robust to their proposed logical perturbations.


\section{Inductive Reasoning}
\label{sec:inductive}

\subsection{Existing Task Formulations}

Existing tasks for inductive reasoning can be summarized as rule verification and rule generation tasks.
Datasets for the tasks are summarized in Table~\ref{tab:inductive_datasets}.
``Generation'' tab with \xmark means it is for the rule verification task.

\paragraph{Rule Verification}
Given a generated $rule$ and $facts$ where the $rule$ is generated from, the task is to classify whether the $rule$ can be accepted. 
The current evaluation aspects are from requirements of both inductive reasoning and natural language.

\paragraph{Rule Generation}
Given multiple manually selected $facts$ with similar patterns, the task is to induce a $rule$ that (1) can entail the $facts$, and (2) is more general than all of the $facts$.
Here ``more general'' means larger information coverage scope. 
More detailed illustrations can be found in \S\ref{appen:general_inductive_reasoning}.

\paragraph{Scientific Hypotheses Generation}
This task is similar to {\it Rule Generation} task but is much more challenging in that the generated $rule$ should not be commonsense knowledge but scientific hypotheses that are even new to humanity.




\begin{table}[]
\resizebox{1.0\columnwidth}{!}{
\begin{tabular}{ccccccccc}
\toprule
Dataset& \thead{Human\\ written} & \thead{Human\\ labeled} & Realistic & \thead{Rule\\ provided} & \thead{Not restricted\\ rule types} & Generation &  \thead{Novel scientific\\ hypotheses}  & Size \\ \midrule
property-norm & \xmark    & \xmark    & \checkmark      & \xmark    & \xmark& \xmark & \xmark &23k  \\
DEERLET& \xmark    & \checkmark   & \checkmark      & \checkmark   & \checkmark      & \xmark & \xmark & 846  \\ \midrule
DEER   & \checkmark   & \checkmark   & \checkmark      & \checkmark   & \checkmark      & \checkmark & \xmark & 1.2k \\ 
ARC  & \xmark &  \xmark  &  -  &  \xmark  &  \xmark  &  -  &  \xmark &  1k \\
\midrule
OpenD5 & \checkmark  &  \checkmark   &  \checkmark   &  \checkmark  &  \checkmark   &  \checkmark  &  - & 675 \\
C-LBD & \checkmark  &  \xmark  &  \checkmark  & \checkmark  &  \checkmark  &  \checkmark & \checkmark &  67k \\
TOMATO & \checkmark  &  \checkmark  &  \checkmark  & \checkmark  &  \checkmark  &  \checkmark & \checkmark & 50 \\ 
\bottomrule
\end{tabular}}
\caption{Summary of inductive reasoning datasets: property-norm~\citep{DBLP:journals/corr/abs-2205-06910}, DEERLET and DEER~\citep{DBLP:journals/corr/abs-2212-10923}, ARC~\citep{DBLP:journals/corr/abs-1911-01547}, OpenD5~\citep{DBLP:journals/corr/abs-2302-14233}, C-LBD~\citep{DBLP:journals/corr/abs-2305-14259}, and TOMATO~\citep{DBLP:journals/corr/abs-2309-02726}.
``Not restricted rule types'' means whether the data is not restricted in a specific topic~(e.g., taxonomic).
}
\label{tab:inductive_datasets}
\end{table}
\subsection{Methods}
{\it Rule Generation} methods almost always have a {\it Rule Verification} step after the initial generation of rules.
To have a clearer overview, we separately introduce the framing or methods of the two tasks.

\subsubsection{Rule Verification}
\label{sec:rule_classification}

\citet{DBLP:journals/corr/abs-2212-10923} propose three requirements of rule verification on inductive reasoning from philosophy literature~($rule$ and $facts$ should not be in conflict; $rule$ should reflect reality; $rule$ should generalize over $facts$) and one requirement of rule verification from NLP requirement~($rule$ should not be trivial or incomplete). 
They focus on inducing $rule$ of many disciplines~(e.g., zoology and history) from $facts$ as textual observations~(e.g. Wikipedia). 
They implement the verification by LLMs~(framing as classification problems).

Another group of works'~\citep{zhu2023large,DBLP:journals/corr/abs-2309-05660,Qiu2023phenomenal} adopted rule verification criteria is compliant with one of the key requirements proposed by \citet{DBLP:journals/corr/abs-2212-10923}, which is that $rule$ and $facts$ should not be in conflict.
They focus on inducing (executable) $rule$ from synthetic $facts$ such as a sequence of number~(example $rule$: find the smallest number), arithmetic calculation~(example $rule$: ``6+4=10''), or changes of 2D grid images~(example $rule$: executable code for moving the grids).
They verify rules by checking the consistency of the labels of annotated examples~($facts$) and the results of $rules$.


\subsubsection{Rule Generation}

\citet{DBLP:journals/corr/abs-2212-10923} 
assume that the inductive reasoning task is so difficult that a proper system should contain a rule populator and (multiple) rule verifiers that filter bad rules from different aspects.
Accordingly, they propose a framework named chain-of-language-models~(CoLM), where one LLM generates $rules$ given $facts$, the other four LLMs filter generated rules mainly based on philosophical requirements of inductive reasoning.

Besides the rule generation and filtering process,
\citet{zhu2023large} further propose to generate rules based on chain-of-thought prompting, and verify rules based on whether the rules can be used to deduce the annotated answer correctly;
\citet{DBLP:journals/corr/abs-2309-05660} further propose that under synthetic datasets, executable code can be generated for the textual rules and verify the rules by executing the code and comparing the results with groundtruth annotation;
 \citet{Qiu2023phenomenal} further propose a third stage of ``rule refinement,''~(leveraging feedback and generate again) and that iteratively repeating the three stages can obtain better rules.

\subsubsection{Scientific Hypotheses Generation}

\citet{DBLP:journals/corr/abs-2302-14233} focuses on proposing hypotheses~(from many disciplines) from a research goal and two comparable corpora. 
Their method also follows a generate-filter process, where LLMs are used for the filtering stage.
\citet{DBLP:journals/corr/abs-2305-14259} focus on proposing NLP hypotheses from a seed term and background context.
Before the hypotheses generation module, they build knowledge graphs to associate academic terms, and retrieve some of the terms as inspirations.
\citet{DBLP:journals/corr/abs-2309-02726} focuses on proposing social science and business hypotheses only from a pile of raw web corpora.
To utilize raw web corpora, they expand generate-filter modules with a background finder module and an inspiration finder module. 
They also propose three feedback mechanisms named past feedback, present feedback, and future feedback to help the inter-communications between modules to induce more novel, valid, and helpful hypotheses.

\section{Abductive Reasoning}
\label{sec:abductive}

\subsection{Existing Task Formulations}

\begin{table}[]
\resizebox{1.0\columnwidth}{!}{
\begin{tabular}{c|cccccc}
\toprule
Dataset & \thead{Human\\ written} & Realistic & Multi-step & \thead{Theory\\ included} & Generation & Size \\ \midrule
$\alpha$NLI    & \checkmark            & 	\checkmark       & \xmark   & \xmark          & \xmark &  22k \\ \midrule
$\alpha$NLG    & \checkmark            & \checkmark       & \xmark   & \xmark          & \checkmark &  76k   \\ \midrule
AbductionRules       & \xmark & \xmark        & \xmark  & \checkmark         & \checkmark  & 114k  \\
D*-Ab   & \xmark & \xmark        & \checkmark   & \checkmark         & \checkmark &  14k   \\
 \bottomrule
\end{tabular}
}

\caption{Summary of abductive reasoning datasets: $\alpha$NLI and $\alpha$NLG~\citep{DBLP:conf/iclr/BhagavatulaBMSH20},
AbductionRules~\citep{DBLP:conf/acl/Young0BW22}, and D*-Ab~\citep{DBLP:conf/acl/TafjordDC21}.
``Realistic'' means whether the data is consistent with the real world.
``Multi-step'' means whether multiple reasoning steps are needed to get the result.
}

\label{tab:abduction_datasets}

\end{table}

Existing tasks for abductive reasoning can be summarized as explanation classification, and explanation generation w/o and w/ theory.
Datasets for the tasks are summarized in Table~\ref{tab:abduction_datasets}.
In the table, the ``generation'' tab and ``theory included'' tab can be used to determine the task it is used for.

\paragraph{Explanation Classification}
Given observation $O_{1}$ at time $t_1$, observation $O_2$ at time $t_2$ ($t_2 > t_1$), a plausible hypothesis $h^+$ and a implausible hypothesis $h^-$ that explain $O_1$ and $O_2$, 
this task
is to select the most plausible hypothesis from $h^+$ and $h^-$. $O_{1}$ and $O_{2}$ each contains a single sentence.

\paragraph{Explanation Generation without Theory}
Given observation $O_{1}$ at time $t_1$, observation $O_2$ at time $t_2$ ($t_2 > t_1$), 
this task
is to generate a valid hypothesis $h^+$ given $O_{1}$ and $O_{2}$. $O_{1}$ and $O_{2}$ each is described in a single sentence.

\paragraph{Explanation Generation with Theory}
Given a theory $C$ and a possible observation $O$ not provable from $C$, the task is to generate a new hypothetical fact $h$ such that $C \cup \{h\} \models O$. 
Here $C$ contains multiple facts and rules, where each fact or rule contains a single sentence. 
$O$ is in single sentence.


\subsection{Methods}

\subsubsection{Explanation Classification}
Methods for this task generally introduce knowledge in various ways to improve performance.
Specifically,~\citet{DBLP:journals/corr/abs-1909-08855} explore ways to incorporate additional unstructured textual knowledge retrieved from a story corpus through prompt;
\citet{DBLP:conf/emnlp/PaulF20} encode and incorporate knowledge from COMET's generation~\citep{DBLP:conf/acl/BosselutRSMCC19} directly into transformer's internal attention;
\citet{DBLP:conf/aaai/LourieBBC21} and~\citet{DBLP:conf/starsem/PaulF21} incorporate knowledge by multi-task training;
%
\citet{DBLP:conf/acl/DuD0Q20} incorporate knowledge with an additional pre-training stage using $\mathcal{ARI}$ independent story corpora;

In addition to knowledge integration, many different aspects of explanation classification tasks are also investigated.
Specifically,
\citet{DBLP:conf/iclr/BhagavatulaBMSH20} rewrite the objective using Bayes Rule 
and formulate a set of probabilistic models that make various independence assumptions on the new objective.
They find that the most sophisticated probabilistic model works the best;
\citet{DBLP:conf/sigir/ZhuPLC20} frame this task as a ranking task to also measure the plausibility of hypothesis in addition to discriminating it;
\citet{DBLP:conf/starsem/PaulF21} conduct this task in an unsupervised setting 
by pretraining on a counterfactual reasoning dataset, which is related to abductive reasoning.
%
\citet{DBLP:conf/naacl/KadikisSK22} propose a method to select suitable LLMs for this task.
It is based on the cosine similarity of $embed(O_1, O_2)$ and $embed(h_i)$ for each LLM without finetuning.
\citet{zhao2023abductive} assume that different $h$ are mutually exclusive, and improve performance by incorporating an additional loss item as regularization to enforce an unbalanced probability prediction over different $h$. 
\citet{chan2023self} exploit inter-sentential coherence and the model consistency to develop a prompt tuning model.

\subsubsection{Explanation Generation without Theory}
In general, methods for this task either incorporate knowledge or improve the decoding method to be more suitable for this task.

For knowledge integration,
\citet{DBLP:conf/iclr/BhagavatulaBMSH20} utilize textual knowledge generated from COMET and investigate two ways of knowledge integration --- via texts or via embeddings, and find that the embedding-based method is more effective;
\citet{DBLP:conf/emnlp/JiKHWZH20} leverage structural knowledge from ConceptNet~\citep{speer2017conceptnet} for this task.

For improving decoding method, 
\citet{DBLP:conf/emnlp/QinSWBHBBC20} are motivated by the fact that the target $h^+$ to generate happens before $O_2$. 
They accordingly propose an unsupervised decoding algorithm that can incorporate both past and future contexts.




\subsubsection{Explanation Generation with Theory}
\citet{DBLP:conf/acl/TafjordDC21} explore the ability of a finetuned T5-11B~\citep{raffel2020exploring} on $P(h|C,O)$. 
Their results indicate that finetuned T5-11B can reach a high test accuracy of 93\% on D*-Ab.


\section{Challenges and Opportunities}
\label{sec:challenges}

We list more challenges and opportunities in \S\ref{appen:other_challenges}.

\paragraph{Computationally Efficient Reasoner}

Many tasks in logical reasoning over formal language have very high algorithmic complexity~\citep{DBLP:journals/ml/MuggletonRPBFIS12}.
Thanks to the low computational cost of each deduction step over formal language, such complex tasks could be possible.
However, each deduction step in LRNLP typically costs one inference of an LLM, which makes tasks with high algorithmic complexity nearly prohibitive.

\paragraph{Robust Deductive Reasoner}
Symbolic deductive reasoners are not restricted to training data distributions, while neural deductive reasoners are restricted to their training data~\citep{DBLP:conf/nips/GontierSRP20,DBLP:conf/aaai/0001S22}.
In addition, neural deductive reasoners are also vulnerable to adversarial attacks~\citep{DBLP:conf/ijcai/GaskellMTS22}, while symbolic reasoners are robust to the attacks.
The lack of robustness can lead to restricted application domains and incorrect deductive inferences.

\paragraph{Better Automatic Evaluation Metrics}

It is generally difficult to automatically evaluate generative reasoning implications, especially with realistic and not synthetic datasets.
The difficulty mainly lies in that the same semantic meaning can be expressed with diversified forms, and that different conclusions might be all acceptable~(especially in abductive and inductive reasoning). 
This may lead to biased evaluation when using automatic metrics.

\paragraph{More Impacts on (NLP) Applications}

As illustrated in \S\ref{sec:advantages_of_new_paradigm}, 
overall LRNLP can be seen as a new type of neuro-symbolic method, which
takes the advantages from both the symbolic and sub-symbolic aspects.
%
These characteristics make an LRNLP system possible (but might still be challenging) to deal with many (NLP) applications such as medical diagnosis and legal NLP tasks,
since many medical and legal problems could be seen as pure logical reasoning problems with very large rule bases~(e.g., medical knowledge and laws).


\paragraph{Probabilistic Inference}

In reality, pure deductive reasoning has not always been used. 
When people include ``likely'' in their expressions, uncertainty is introduced, which makes the reasoning process probabilistic;
in addition, inductive reasoning and abductive reasoning are by default non-monotonic reasoning.
This uncertainty aspect has not been focused in current research. 
It is probably beneficial to learn from how symbolic reasoning handles uncertainty~\citep{halpern2017reasoning}.

\paragraph{Reasoning with Incomplete Information}

The current proof generation task requires all necessary premises provided to create a proof tree.
Only one work~\citep{DBLP:journals/corr/abs-2211-00614} focuses on proof generation with the incomplete information task. 
However, the task they adopt 
only overlooks
one premise, while in reality more might be missing.

\paragraph{Inductive Reasoning on Web Corpora}

Currently, the dataset for rule generation tasks in inductive reasoning provides manually selected facts~\citep{DBLP:journals/corr/abs-2212-10923}.
However, to best leverage a system's ability to handle natural language, it should be able to work on raw web corpora to induce rules, which leads to a more challenging task of inductive reasoning on web corpora.

\paragraph{Abductive Reasoning with (Long) Theory}

Many tasks such as medical diagnosis conduct abductive reasoning with a long theory~(e.g., medical knowledge).
However, current abductive reasoning research only covers abductive commonsense reasoning~\citep{DBLP:conf/iclr/BhagavatulaBMSH20} without given theory, or only given short, synthetic, not realistic knowledge as theory~\citep{DBLP:conf/acl/TafjordDC21}.

\paragraph{Interactions between Reasoning Types}

Multiple reasoning types can be used together for complex tasks.
Existing works only utilize deductive reasoning with abductive reasoning to create a proof tree~\citep{DBLP:conf/naacl/HongZYZ22,DBLP:journals/corr/abs-2211-00614}.
However, many other collaborations are possible, such as using inductive reasoning to collect a (large) rule base, which is to be used as the theory base for deductive reasoning.

\section{Conclusion}
\label{sec:conclusion}

In this survey, we review papers using transformer-based LLMs explicitly working on deductive, inductive, and abductive reasoning over English representation.
Specifically,
we have introduced the philosophical foundations, advantages of LRNLP, benchmarks and methods, challenges of LRNLP, possible future directions, and the relation of LRNLP to related NLP fields~(\S\ref{appen:relation_to_related_fields}).

\section*{Limitations}
In consideration of space constraints, this paper focuses more on (1) providing a high-level overview and prospect of the LRNLP field~(e.g., advantages and challenges of the field), and (2) delineating the broader evolutionary trajectories of pertinent methodologies.
It might not include all the details of the surveyed papers.

\section*{Ethics Statement}
This article follows the ACL Code of Ethics. 
To our knowledge, there are no foreseeable potential risks to use the datasets and methods in this paper.

\bibliography{custom}

\appendix

\section{Appendix}
\label{sec:appendix}

\subsection{Relation to Related (NLP) Fields}
\label{appen:relation_to_related_fields}
In this section, we first introduce related NLP fields to general logical reasoning, then introduce fields that are only related to deductive reasoning, inductive reasoning, or abductive reasoning. 
We hope that this section could be helpful to form a clear shape of LRNLP in NLP.

\subsubsection{Logical Reasoning}
There are some previous works involve the term ``logical reasoning'', but do not provide a specification on which sub-type of logical reasoning they involve.
In many cases these works are more close to ``natural language inference'', which adopts datasets where the data involve a mixture of multiple sub-types of logical reasoning, making it hard to analyze from each sub-type. Therefore we do not include these works in this survey.

\paragraph{Neuro-Symbolic Computing}
Neural-symbolic computing is a hybrid of symbolism and connectionism to exploit advantages from both sides~\citep{wang2022towards,DBLP:conf/lrec/CambriaLDXK22}.
The knowledge representation of its symbolic part basically is a knowledge graph or propositional logic or first-order logic~\citep{wang2022towards}.
LRNLP could be seen as a new type of neuro-symbolic in addition to the existing 6 types summarized by ~\citet{kautz2022third}, as its goal and design of methodology are typically symbolic~(logical reasoning with knowledge bases), while avoiding any symbolic representation, using~(currently pure) neural methods.


\paragraph{Natural Language Inference}
Natural language inference~(NLI) is generally considered as the semantic concepts of entailment and contradiction~\citep{DBLP:conf/emnlp/BowmanAPM15}.
Here logical reasoning tasks can be viewed as special types of NLI focusing on particular reasoning aspects. 

\paragraph{Question Answering}
The form of LRNLP looks similar to question answering~(QA), however, QA is conducting one-step logical reasoning only when the context provides enough information to answer the question~(deductive reasoning), or the answer is a generalization of an argument in context or question~(inductive reasoning), or the answer is to provide explanations to the question~(abductive reasoning).

\paragraph{Commonsense Reasoning}
Commonsense reasoning~(CR) and logical reasoning~(LR) are similar in that they both involve ``knowledge'' and ``reasoning''.
Compared to LR, CR focuses more on the ``knowledge'' aspect.
Some typical tasks include whether a system has commonsense knowledge~\citep{DBLP:conf/acl/BosselutRSMCC19,DBLP:conf/emnlp/YangDRC20}, and whether a system's answer is commonsense-knowledge-aware~\citep{DBLP:conf/aaai/BiskZLGC20};
LR focuses more on the ``reasoning'' aspect, e.g., whether a system's i/o behaviors follow reasoning requirements~\citep{DBLP:conf/ijcai/ClarkTR20}.

\paragraph{Chain of Thoughts}
\label{subsec:chain-of-thoughts}
Chain of thoughts~(COT)~\citep{DBLP:journals/corr/abs-2201-11903} is a prompting technique that can elicit the step-by-step reasoning ability of LLMs without finetuning.

COT can potentially be used for each of the three sub-reasoning types of logical reasoning.
In fact, for a given~(commonsense reasoning) question, some reasoning steps of COT could be deductive, and others can be inductive or abductive. 
Since the purpose of this paper is to provide a finer analysis on logical reasoning, we do not intentionally cover prompting techniques such as COT.

It is also argued by several modular-based deductive reasoning methods that COT's reasoning is not causal~\citep{DBLP:journals/corr/abs-2205-09712}, limited by input size~\citep{DBLP:conf/naacl/Ribeiro0MDWZCXH22}, and contains unrelated or incorrect steps~\citep{DBLP:conf/naacl/HongZYZ22,DBLP:journals/corr/abs-2210-12217}.

Overall, it could be interesting to use COT-related methods specifically for deductive, inductive, or abductive reasoning~(as opposed to modular-based methods), and it is a less-explored research direction.

\subsubsection{Deductive Reasoning}
\label{append:relation_deductive_reasoning}

\paragraph{Multi-hop Reasoning}
Compared to proof generation, many multi-hop reasoning tasks~\citep{DBLP:conf/emnlp/Yang0ZBCSM18,DBLP:conf/emnlp/JiangBZD0B20,DBLP:conf/acl/MinZZH19,DBLP:conf/emnlp/SinhaSDPH19} are much simpler, often being single-branched~\citep{DBLP:conf/naacl/QuCGDX22}, consisting of only 2-3 supporting facts, and are more coarse-grained, involving large chunks of texts such as passages instead of simple, short sentences~\citep{DBLP:journals/corr/abs-2205-12443}. 

Nevertheless, some multi-hop reasoning datasets can be considerd as conducting deductive reasoning.
For instance, for each data in CLUTRR~\citep{DBLP:conf/emnlp/SinhaSDPH19} dataset, a set of facts that can make conclusive support to the target kinship relation is included in background information as input for each target relation, hence from the philosophical definition~\citep{salmon1989introduction}, it requires to perform deductive reasoning.


\paragraph{Mathematical Reasoning}
In many mathematical reasoning tasks such as math word problem solving~\citep{DBLP:journals/tacl/Koncel-Kedziorski15} and geometry problem solving~\citep{DBLP:conf/emnlp/SeoHFEM15},
the conclusion can be conclusively entailed by the premise.
Therefore these tasks belong to deductive reasoning.
We do not review math-related papers because we want to focus solely on the challenge of deductive reasoning while mathematical reasoning involves numbers in the text, which introduces additional challenges.

\subsubsection{Inductive Reasoning}

\paragraph{Information Extraction}
Information Extraction~(IE) is a task of extracting pre-specified types of facts from written texts or speech transcripts, and converting them into structured representations~\citep{DBLP:reference/db/Ji18}.
The rule generation task here also extracts rules from facts represented in written texts.
The difference is that IE pursues extracting the exact information from existing texts, while inductive reasoning aspires to induce more general rules from existing texts, where the information in rules goes beyond what is exactly stated in the texts.

\paragraph{Case-based Reasoning}
Case-based Reasoning~(CBR) is a classic AI subject, whose methods share a general methodology of four steps: retrieve, reuse, revise, and retain~\citep{aamodt1994case}.
Recently there has been research works devoting to bridging the research of CBR and NLP, by using NLP techniques for CBR challenges~\cite{yang-etal-2023-end} and improving NLP tasks with CBR methodologies~\cite{DBLP:conf/emnlp/DasZTGPLTPM21,DBLP:conf/icml/DasGNTZHJM22,yang-etal-2023-end,thai2023machine}.
CBR could be seen as a type of analogical reasoning~\citep{kolodner1997educational}, and analogical reasoning belongs to inductive reasoning~\citep{salmon1989introduction}.
However, CBR is a different inductive reasoning type than the ``generalization'' process~(from facts to rules) described in~\citet{flach2000abductive}, but more on the general description on inductive reasoning~\citep{salmon1989introduction} that premises cannot conclusively provide support to the conclusion.

\subsubsection{Abductive Reasoning}

\paragraph{Causal Reasoning}
In logic research, causal reasoning aims at an epistemological problem of establishing precise causal relationships between causes and effects. 
It is generally considered a form of inductive reasoning~\citep{goertzel2011real}, since inductive reasoning is to derive rules that lead from one to another.
When the focus is to derive possible causes from effects, the problem belongs to abductive reasoning~\citep{goertzel2011real}.



\subsection{Full Details About the Definition and Categorization of Logical Reasoning}
\label{appen:difference_inductive_abductive}

There are many subjects related to logical reasoning, including philosophy, logic, and AI.
Among them, the definition and categorization aspects of logical reasoning are handled by philosophy research. 
However, debate exists in philosophy research on the categorization of logical reasoning.

One group believes that every argument can be classified as either deduction argument, inductive argument, or fallacy~\citep{salmon1989introduction}. 
Without considering fallacy, given that an argument consists of premises and a conclusion, when the premises can conclusively provide support to the conclusion~(which means that if the premises of the argument were all true, it would be impossible for the conclusion of the argument to be false), this argument is a deductive argument.
Conversely, when the premises can not conclusively provide support to the conclusion, the argument is 
inductive.

The other group has the same definition of deductive reasoning, but they believe that further categorization of non-deductive reasoning is necessary.
Without considering fallacy, they believe in a trichotomy of deductive, inductive, and abductive reasoning~\citep{peirce1974collected}.
However, even for the second group, the definition and difference between inductive and abductive reasoning are also controversy~\citep{flach2000abductive}. 

Nevertheless,~\citet{console2000relations} argue that from the utility perspective of AI, a distinction between inductive and abductive reasoning is possible: both inductive and abductive reasoning provide explanations about the world but their explanations differ in the degree of generality.
For instance, an inductive hypothesis allows the validity of properties, observed on a set of individuals, to be generalized to other individuals not in the observations, whereas an abductive one allows unobserved properties to be applied to observed individuals.

Considering that inductive and abductive reasoning can be distinctive enough when formulated in NLP,
in this paper, we adopt the second group, particularly~\citet{console2000relations}'s view of definition and categorization of logical reasoning. 

Specifically, the difference between inductive and abductive reasoning is that, both inductive and abductive reasoning provide explanations about the world but their explanations differ in the degree of generality.

For instance, an inductive hypothesis allows the validity of properties, observed on a set of individuals, to be generalized to other individuals not in the observations, whereas an abductive one allows unobserved properties to be applied to observed individuals.

The distinction between inductive and abductive hypotheses strictly parallels the dichotomy {\it extension} vs. {\it intension}, or {\it generality} vs. {\it informativeness}. 
In other words, an inductive hypothesis extends or generalizes to unobserved individuals, while an abductive one provides more specific information (e.g., unobserved properties) about existing specific individuals.

For example, if a white ball is found in a bag, inductive reasoning might lead to the conclusion that ``all balls in this bag are white'', while abductive reasoning might lead to the conclusion that ``someone put the white ball into this bag''. 

In this example, the inductive hypothesis generalizes the property of the existing individual~(the white ball) to unobserved individuals~(other not-seen balls in the bag), while the abductive hypothesis provides more specific information about the current individual~(who put this ball to the bag).

To summarize in simple words, in common situations, pure inductive reasoning is to only provide (usually sample to population) generalizations, while pure abductive reasoning is to only provide specific explanations.


Overall, even in the philosophical literature (which takes charge of the research on the definition of logical reasoning), a clear definition for all three types of logical reasoning is rare, but more on the description of the difference between types of logical reasoning (since a clear definition is still under debate). The difference can be illustrated does not mean a precise definition can be given. Nevertheless, considering the above-discussed philosophical literature, we try our best to give a definition below for a more straightforward understanding:

Given an argument consisting of premises and a conclusion, we define the sub-type of logical reasoning it involves below:

{\it Definition for deductive reasoning}: the premises can conclusively provide support for the conclusion, i.e. if the premises are all true, it would be impossible for the conclusion to be false.

{\it Definition for inductive reasoning}: the premises cannot conclusively provide support for the conclusion, since the conclusion generalizes existing information in premises to new knowledge, which has a wider applicable scope than those in premises.

{\it Definition for abductive reasoning}: the premises cannot conclusively provide support for the conclusion, since the conclusion contains more specific information over the premises~(most commonly used as generating most probable explanations).

Please note that according to \citet{console2000relations}, inductive reasoning and abductive reasoning are not exclusive to each other, i.e., inductive reasoning and abductive reasoning overlap with each other.

\subsection{Why We Choose Definition in Section \ref{sec:definition}}
\label{appen:why_choose_this_definition}

Firstly, some other definitions~(e.g., from Pieces) are not in conflict with the one we adopted. Secondly, other definitions lack a clear boundary between different types of reasoning, while our adopted definitions clearly delineate such boundaries (e.g., general vs. specific for inductive and abductive reasoning).

To elaborate why there's no contradiction, specifically, Pierce's definition is "inference to the most plausible explanation for incomplete observations". Here "explanation" refers to not guaranteed and specific information. An example about the discussion on "specific" can be found in \S\ref{appen:general_inductive_reasoning}. We use definition in \S\ref{sec:definition} because other definitions lack a clear boundary between deductive, inductive, and abductive reasoning, while our adopted definitions clearly delineate such boundaries (e.g., general vs. specific for inductive and abductive reasoning; guaranteed vs. not guaranteed for deductive and the remaining two reasoning).

\subsection{Related Surveys on Reasoning}
\label{appen:related_reasoning_surveys}

\citet{DBLP:journals/corr/abs-2212-10403,DBLP:journals/corr/abs-2212-09597} mainly reviews the prompting techniques for LLMs, but do not focus on papers that specialized on logical reasoning~(the coverage of the two fields are quite different).
\citet{yu2023nature} also review papers related to reasoning.
However, 
(1) they do not focus on logical reasoning, and do not organize their survey based on the three sub-types of logical reasoning. Particularly, only a small section discusses this topic;
(2) their definition on the deductive, inductive, and abductive reasoning lacks a philosophy foundation~(no reference), and is confusing. 
Particularly, from their definition, it is unclear on the difference between inductive and abductive reasoning.
Specifically, it is unclear on what is the difference between the ``more general rule'' and ``best explanation''? 
A more general rule, such as Newton's Law, can also serve as the best explanation about phenomenons related to object movement. 
In the contrary, this survey's definition is based on philosophy literature~\citep{console2000relations}, and our definition can clearly differ between inductive and abductive reasoning. 
The difference lies in that inductive reasoning is about ``general'', and abductive reasoning is about ``specific'', while a specific conclusion, such as ``it must have rained since the lawn is wet'', is commonly used as ``the best explanation''.
But the point of abductive reasoning is about ``specific'', not ``explanation'', since inductive reasoning can also provide explanation~\citep{flach2000abductive}.
We illustrate in \S\ref{sec:definition} that there has been various forms of definition for the three reasoning types during the thousands of years of development of the philosophy research. 
The variance of definition should be aware and the definitions should be given with a systematic view of the philosophy research to avoid confusion.
We provide a detailed discussion about the categorization of logical reasoning from a philosophy perspective in \S~\ref{appen:difference_inductive_abductive}.

\citet{yu2023nature} do not stress and organize the survey from the three sub-types of logical reasoning.
\citet{xu2023large} provides a comprehensive evaluation of the logical reasoning ability of LLMs. 
They are not to provide a survey but to use LLMs on the existing logical reasoning datasets.

\citet{lakoff1970linguistics,mccarthy1990example} are the first few works to take a close look at the connection between logical reasoning and natural language.
\citet{dagan2005pascal} proposed evaluating logical reasoning through the comparison of two natural language texts.
\citet{maccartney2014natural} apply logical reasoning in natural language inference through iterative editing of natural language.

\subsection{Other Inductive Reasoning Papers}
\paragraph{Implicit Rule Verification}
\citet{DBLP:journals/corr/abs-2205-06910} analyze language model's ability to generalize novel property knowledge~(has sesamoid bones) from concept(s)~(robins) to others~(sparrows, canaries).
As illustrated in \S\ref{appen:general_inductive_reasoning}, they analyze the language models' ability to classify a new fact~(but not a rule) as correct or not, given facts.
It could be seen that the correctness of a rule is implicitly predicted by testing multiple facts entailed by the rule.

\paragraph{Symbolic Rule Generation}
\citet{DBLP:conf/acl/WangP22} propose attentive memories with novel
differentiable logic operators to induce symbolic rules from texts.

\subsection{Research Trend in the Three Sub-Types of Logical Reasoning}
\label{appen:trend}
Out of the three reasoning types, deductive reasoning has drawn the most research attention, and has the most abundant of works, especially in 2022. Abductive reasoning has drawn much attention in 2020 and 2021 but has few works in 2022 and 2023. Inductive reasoning is only proposed at the end of 2022, having the least number of works. However, inductive reasoning has attracted much attention since the second half year of 2023.

Two main reasons for the abundance of works in the deductive reasoning domain could be that (1) more challenging benchmarks have been constructed during the last few years, and (2) deductive reasoning could be one of the most commonly used reasoning types in common life. We think the main reason for the little attention drawn to abductive reasoning in recent years is that the benchmarks for abductive reasoning are relatively old and less challenging for LLMs. Inductive reasoning could be a promising research topic since there have been few works in the domain, and it involves very challenging tasks such as proposing new scientific findings.

In general, there has been no framework which is proposed to address all three reasoning domains. However, LLMs generally can exhibit all three reasoning abilities to some extent. It would be interesting for future works to analyze the effect of the pretraining method and scale of LLM on the three reasoning abilities.

\subsection{Relation Between LRNLP and neuro-symbolic}

A large proportion of recent papers on deductive reasoning and abductive reasoning leverage a natural language-based knowledge base, and reason over retrieved knowledge from the knowledge base to reach a certain goal~\citep{DBLP:conf/acl/TafjordDC21,DBLP:conf/naacl/LiangBS21,DBLP:conf/naacl/QuCGDX22,DBLP:conf/naacl/Ribeiro0MDWZCXH22,DBLP:journals/corr/abs-2205-09712,DBLP:conf/acl/SanyalS022,DBLP:conf/naacl/HongZYZ22,DBLP:journals/corr/abs-2201-06028,DBLP:journals/corr/abs-2205-12443,DBLP:journals/corr/abs-2210-12217,DBLP:journals/corr/abs-2204-13074}. This pattern is very similar to the methodology design of neuro-symbolic methods, which is to retrieve symbolic knowledge and reason over the retrieved symbolic knowledge. The main difference is that LRNLP adopts natural language as knowledge representation but not symbolic knowledge. Because of the similarity in the methodology design, we consider that LRNLP could be seen as a type of neuro-symbolic methods but without many disadvantages of symbolic representation such as symbolic knowledge acquisition and scalability.

In addition, due to the high similarity in the methodology design to neuro-symbolic, LRNLP also shares some advantages with neuro-symbolic such as explainability. The reason is that the iterative retrieving and reasoning will make the decision-making process more interpretable on the intermediate reasoning steps, and which knowledge is used for each reasoning step.

\begin{table*}[]
\centering
\resizebox{2.0\columnwidth}{!}{
\begin{tabular}{c|c|c|ccccccc|c|c}
\toprule
\multirow{2}{*}{Methods} & \multicolumn{1}{c|}{ParaRules}          & \multicolumn{1}{c|}{Birds-Electricity}  & \multicolumn{7}{c|}{EntailmentBank (Task 3)}                                                                                                                                                                                                                           & \multicolumn{1}{c|}{OBQA}     & \multicolumn{1}{c}{QuaRTz}   \\ 
                         & \multicolumn{1}{c|}{Full Accuracy (FA)} & \multicolumn{1}{c|}{Full Accuracy (FA)} & \multicolumn{1}{c}{Leaves F1} & \multicolumn{1}{c}{Leaves All-Cor.} & \multicolumn{1}{c}{Steps F1} & \multicolumn{1}{c}{Steps All-Cor.} & \multicolumn{1}{c}{Intermediates F1} & \multicolumn{1}{c}{Intermediates All-Cor.} & \multicolumn{1}{c|}{Overall All-Correct} & \multicolumn{1}{c|}{Accuracy} & \multicolumn{1}{c}{Accuracy} \\ \midrule
PRover                   & 95.1                                   & 80.5                                   & -                             & -                                   & -                            & -                                  & -                                    & -                                          & -                                       & -                            & -                            \\
multiPRover              & 94.5                                   & 81.8                                   & -                             & -                                   & -                            & -                                  & -                                    & -                                          & -                                       & -                            & -                            \\
EntailmentWriter         & -                                      & -                                      & 39.7                          & 3.8                                 & 7.8                          & 2.9                                & 36.4                                 & 13.2                                       & 2.9                                     & -                            & -                            \\
ProofWriter              & 98.5               & 97.0                                   & -                             & -                                   & -                            & -                                  & -                                    & -                                          & -                                       & -                            & -                            \\
EVR                      & -                                      & 63.1                                   & -                             & -                                   & -                            & -                                  & -                                    & -                                          & -                                       & -                            & -                            \\
IBR                      & 95.7                                   & 93.5                                   & -                             & -                                   & -                            & -                                  & -                                    & -                                          & -                                       & -                            & -                            \\
IRGR                     & -                                      & -                                      & 45.6                          & 12.1                                & 16.3                         & 11.8                               & 38.8                                 & 36.5                                       & 11.8                                    & -                            & -                            \\
Selection-Inference      & -                                      & -                                      & -                             & -                                   & -                            & -                                  & -                                    & -                                          & -                                       & -                            & -                            \\
FaiRR                    & 98.6                                   & -                                      & -                             & -                                   & -                            & -                                  & -                                    & -                                          & -                                       & -                            & -                            \\
MetGen                   & -                                      & -                                      & 34.8                          & 8.7                                 & 9.8                          & 8.6                                & 36.7                                 & 20.4                                       & 8.6                                     & -                            & -                            \\
SCSearch                 & -                                      & -                                      & -                             & -                                   & -                            & -                                  & -                                    & -                                          & -                                       & -                            & -                            \\
ADGV                     & -                                      & -                                      & -                             & -                                   & -                            & -                                  & -                                    & -                                          & -                                       & -                            & -                            \\
NLProofS                 & -                                      & -                                      & 43.2                          & 8.2                                 & 11.2                         & 6.9                                & 42.9                                 & 17.3                                       & 6.9                                     & -                            & -                            \\
Entailer                 & -                                      & -                                      & -                             & -                                   & -                            & -                                  & -                                    & -                                          & -                                       & 76.8                         & 74.3                         \\
Teachme                  & -                                      & -                                      & -                             & -                                   & -                            & -                                  & -                                    & -                                          & -                                       & 77.0                         & 75.9                        \\ \bottomrule
\end{tabular}}
\caption{Proof Generation Task Results.}
\label{tab:deductive_results}
\end{table*}
\subsection{Experiments Summarization}
\label{appen:evaluation}

In this section, we summarize the experiment results of an important and literature-abundant task.

Until now there has been only one or two papers working on inductive reasoning.
Methods for abductive reasoning generally leverage different resources~(such as multi-task, additional knowledge resources, and ancillary loss) and lack a progressive relationship between each other, therefore are less comparable.
Currently, the $Proof Generation$ task in deductive reasoning is the most literature-abundant, and methods for this task have progressive relationships with each other.
Therefore here we mainly summarize results and analyze for the $Proof Generation$ task.

Table~\ref{tab:deductive_results} shows the summarized experiment results. 
We select the most widely used tasks to display their performance.
Among the task, the setting of ParaRules is trained on D3~(D* dataset with depth 3) and tested on the ParaRules test set; the setting of Birds-Electricity is trained on D5~(D* dataset with depth 5) and tested on bird-electricity set; setting for EntailmentBank is the task 3 which uses full corpus as input~(so that many distractors exist in input); setting for OBQA and QuaRTz are zero-shot setting while model pre-trained on another dataset~(EntailmentBank).

Among the methods,~\citet{DBLP:journals/corr/abs-2205-09712} and~\citet{DBLP:journals/corr/abs-2201-06028} design unique metrics using EntailmentBank dataset.
\citet{DBLP:journals/corr/abs-2211-00614} focus on a unique task~(proof generation task with incomplete information), therefore we do not list their experiments results in the table.
Specifically, \citet{DBLP:journals/corr/abs-2205-09712} work on metric of accuracy.
\citet{DBLP:journals/corr/abs-2201-06028} work on metrics of \textit{Goal\%}, \textit{\#Steps}, where \textit{Goal\%} measures the number of valid goals reached by each system, and \textit{\#Steps} measures the number of steps expanded before reaching a valid goal.
\citet{DBLP:journals/corr/abs-2211-00614} work on ``proof generation task with incomplete information'', which naturally performs worse than the ``proof generation task''.

Overall methods for proof generation tasks tend to use different datasets for evaluation, making them less comparable.

\begin{table*}[]
\centering
\resizebox{2.0\columnwidth}{!}{
\begin{tabular}{c|c|c|c|c}
\toprule
                    & Model usage                                                                                                        & \thead{Second-round \\pretraining data \\used before finetuning} & \thead{Experiment \\setting}      & Task \\ \midrule
RuleTakers          & RoBERTa-large~(355M)                                                                                                      & RACE                                                 & finetuning              & 0                                                        \\ \midrule
Leap-of-Thought     & RoBERTa-large                                                                                                      & -                                                    & finetuning              & 0                                                        \\ \midrule
FOLIO               & {[}BERT and RoBERTa{]}-{[}base and large{]}                                                                        & -                                                    & \thead{finetuning \\\& few-shot} & 0                                                        \\ \bottomrule
PRover              & RoBERTa-large                                                                                                      & -                                                    & finetuning              & 1                                                        \\ \midrule
multiPRover         & RoBERTa-large                                                                                                      & -                                                    & finetuning              & 1                                                        \\ \midrule
EntailmentWriter    & T5-11B                                                                                                             & -                                                    & finetuning              & 1                                                        \\ \midrule
ProofWriter         & T5-11B                                                                                                             & -                                                    & finetuning              & 1                                                        \\ \midrule
EVR                 & T5-small~(60M)                                                                                                     & -                                                    & finetuning              & 1                                                        \\ \midrule
IBR                 & RoBERTa-large                                                                                                      & -                                                    & finetuning              & 1                                                        \\ \midrule
IRGR                & \thead{T5-large~(770M) for generation model; \\MPNet-base-v2 for retriever}                                              & -                                                    & finetuning              & 1                                                        \\ \midrule
Selection-Inference & Chinchilla-7B                                                                                                      & -                                                    & \thead{finetuning \\\& few-shot} & 1                                                        \\ \midrule
FaiRR               & \thead{RoBERTa-large for fact and rule selectors;\\ T5-large for knowledge composer}                                         & -                                                    & finetuning              & 1                                                        \\ \midrule
MetGen              & \thead{T5-large for deduction and abduction models; \\ALBERT-xxlarge-v2 for controller}                                      & -                                                    & finetuning              & 1                                                        \\ \midrule
SCSearch            & \thead{T5-large for deduction model; \\DeBERTa-v3-large~(350M) for goal entailment model}                             & \thead{DeBERTa-v3-large\\ on MNLI}                             & finetuning              & 1                                                        \\ \midrule
ADGV                & \thead{T5-large for deduction model; \\T5-3B for abduction model; \\DeBERTa-v3-large for goal entailment model} & \thead{DeBERTa-v3-large\\ on MNLI}                             & finetuning              & 1                                                        \\ \midrule
NLProofS            & \thead{T5-large for deduction model; \\RoBERTa-large for verifier}                                                            & -                                                    & finetuning              & 1                                                        \\ \midrule
Entailer            & T5-11B                                                                                                             & -                                                    & finetuning              & 1                                                        \\ \midrule
Teachme             & T5-11B                                                                                                             & -                                                    & \thead{human \\interaction}       & 1                 \\
\bottomrule
\end{tabular}
}

\caption{A collection of model usage, (second-round) pretraining data usage, and experiment setting for methods in hypothesis classification~(denoted as ``0'' in ``Task'' column) and proof generation~(denoted as ``1'' in ``Task'' column) tasks.}
\label{tab:model_usage_pretraining_data_usage}
\end{table*}
\subsection{Model Structure, Pretraining Data Used, and Experiment Settings}
\label{appen:model_structure_pretraining_data_used}

Table~\ref{tab:model_usage_pretraining_data_usage} shows a collection of model structure, pretraining data usage, and experiment settings for methods in hypothesis classification and proof generation tasks.

In general, RoBERTa-large and T5-11B are the most adopted base models.

\subsection{Meaning of ``More General'' Required by Inductive Reasoning}
\label{appen:general_inductive_reasoning}

This section is collected from~\citet{DBLP:journals/corr/abs-2212-10923}'s appendix, to help illustrate inductive reasoning.

Given an argument consisting of a premise and a conclusion, if the conclusion involves new information that is not covered by the premise and can not be conclusively entailed by the premise, the argument is an inductive argument~\citep{salmon1989introduction}.

When the conclusion has a larger scope of information coverage than the premise, and can entail the premise, it can be said that the conclusion is ``more general'' to the premise~\citep{DBLP:journals/corr/abs-2212-10923}. 
In this case, we termed the premise as a ``fact'', and the conclusion as a ``rule'';
When the conclusion contains new pieces of information and cannot entail the premise, as defined by~\citet{salmon1989introduction}, the argument is still an inductive argument.
But in this case, we termed the premise as a ``fact'', and the conclusion as another ``fact''.

For instance, if facts that are about cats and dogs are good accompaniment of humans, then some examples of a ``more general'' rule can be (1) mammals are good accompaniment of humans, or (2) domesticated animals are good accompaniment of humans, or (3) animals with four legs are good accompaniment of human.

In these examples, the rules cover a larger scope than the facts~(e.g., mammals compared to cats; domesticated animals compared to cats), and therefore the rules are ``more general'' than the facts.

``More general'' means not only about finding higher taxonomic rank, but can be in unlimited forms.
For instance, if the fact is about the Sun rises and falls every day, then some examples of a ``more general'' rule can be (1) the Earth is the king of the universe or (2) the Earth is rotating itself.

Both rule examples are ``more general'' than the given fact, since the rule can entail not only the given fact, but also other not mentioned facts such as the observable movements of the other stars in the Milky Way.

\subsection{Other Challenges and Possible Future Directions}
\label{appen:other_challenges}


\paragraph{Reliable Rule Generation}

Currently, the rule generation method in inductive reasoning relies on out-of-box LLMs, since a finetuned rule generation model could be restricted in a domain.
The annotation of an inductive reasoning dataset should only be done by experts and is very time consuming~\citep{DBLP:journals/corr/abs-2212-10923}.
Given the two restrictions, how to improve the quality of generated rules given related facts could be a challenging open problem.

\paragraph{Reliable Explanation Generation}

Abduction is a form of non-monotonic reasoning~\citep{DBLP:journals/air/Paul93}, and potentially has a large search space of conclusions given premises.
Therefore, how to generate more~(all) reasonable explanations can be challenging~\citep{DBLP:conf/iclr/BhagavatulaBMSH20}.

\paragraph{Building Larger Benchmarks}
For complicated reasoning tasks especially in realistic and natural language settings, usually experts are needed for annotation, and the process is very time-consuming~\citep{DBLP:conf/emnlp/DalviJTXSPC21,DBLP:journals/corr/abs-2211-00614,DBLP:journals/corr/abs-2212-10923}.
Therefore it can be challenging to construct significantly larger benchmarks.

\paragraph{Understanding the Internal Mechanism of LLMs for Reasoning}

Until now research works only focused on investigating whether the input/output behaviors of LLMs can be used to simulate a reasoner~\citep{DBLP:conf/ijcai/ClarkTR20} or complete reasoning tasks.
However, it is still a challenging open research question to understand the internal mechanism of LLMs for reasoning.

\paragraph{Reliable Verifier}
Most current verifiers~(refiners) relies on the internal beliefs of LLMs to select~(improve) from generated rules and mitigate hallucination. 
It's doubtful whether LLMs have obtained the necessary knowledge.

\end{document}